# Efficient Transformations in Deep Learning Convolutional Neural Networks

E4040.2024Fall.PBDY.report

Daniel Harvey
Dept. of Computer Science
School of Engineering and Applied Science
Columbia University
New York, NY USA
dyh2111@columbia.edu

Berk Yilmaz
Dept. of Electrical Engineering
School of Engineering and Applied Science
Columbia University
New York, NY USA
by2385@columbia.edu

Prajit Dhuri
Dept. of Biomedical Engineering
School of Engineering and Applied Science
Columbia University
New York, NY USA
pad2176@columbia.edu

*Abstract*—**This study investigates the integration of signal processing transformations—Fast Fourier Transform (FFT), Walsh-Hadamard Transform (WHT), and Discrete Cosine Transform (DCT) within the ResNet50 convolutional neural network (CNN) model for image classification. The primary objective is to assess the trade-offs between computational efficiency, energy consumption, and classification accuracy during training and inference. Using the CIFAR-100 dataset (100 classes, 60,000 images), experiments demonstrated that incorporating WHT significantly reduced energy consumption while improving accuracy. Specifically, a baseline ResNet50 model achieved a testing accuracy of 66%, consuming an average of 25,606 kJ per model. In contrast, a modified ResNet50 incorporating WHT in the early convolutional layers achieved 74% accuracy, and an enhanced version with WHT applied to both early and late layers achieved 79% accuracy, with an average energy consumption of only 39 kJ per model. These results demonstrate the potential of WHT as a highly efficient and effective approach for energy-constrained CNN applications.**

## I. Introduction

Neural networks have emerged as one of the most significant advancements in computing, capable of approximating any distribution or mathematical function through the Universal Approximation Theorem [1]. However, this capability comes at a significant cost: deep learning models are computationally expensive. A model's input layer, representing the dimensionality of the data, is processed through multiple layers and neurons, often leading to models with millions of parameters. In extreme cases, such as with the Transformer-based GPT-MoE (Mixture of Experts), its current models can scale up to 1.8 trillion parameters [2]. This complexity translates to substantial computing power requirements, leading to higher energy consumption and an increased carbon footprint.

As AI infrastructure scales, large technology companies face growing challenges, with energy availability rather than computational capacity emerging as the primary bottleneck. For instance, Microsoft has adopted long-term energy strategies, including investments in solar farms and nuclear power, to support its AI operations [3]. These efforts underscore the urgent need to optimize AI computations, not only to reduce energy demands and costs but also to mitigate environmental impact.

Reducing the dimensionality of complex data, combined with implementing efficient computations like matrix multiplication, directly lowers both power consumption and computational demands. However, this approach often introduces a performance tradeoff, as lower-dimensional representations may fail to fully capture the rich information embedded in the original data. Balancing this tradeoff—minimizing computational requirements while preserving performance—is especially critical in constrained environments such as self-driving cars and embedded systems, where space and power are limited. In these scenarios, optimizing algorithms and data representations is essential to achieving energy efficiency and hardware compatibility, supporting the broader goal of developing sustainable and scalable AI solutions.

An area of particular interest in high-dimensional data is visual media, including photos and videos. These are widely used as sensing tools in industrial applications such as object recognition, positioning, and security, as well as in consumer applications like video games and mobile devices. Visual data is typically represented in a 2D grid format, with dimensions of width × height and three color channels—red, green, and blue (RGB). When combined additively, these channels create a full-color image.

For instance, a prosumer camera like the Canon R10 captures 24-megapixel images, represented as 6000 × 4000 pixels in RGB format. Flattening this data into a one-dimensional array results in 72 million values. If each value is stored as a 32-bit (4-byte) floating-point number as typically done in the popular machine learning framework TensorFlow, processing a single image requires approximately 288 MB of memory. In contrast, a still image displayed on an HD TV, with a resolution of 1920 × 1080 pixels in RGB, contains only 6.22 million values, requiring about 25 MB of memory—nearly 10 times less. This highlights the challenges of working with high-resolution, high dimensional data, in constrained environments.

When visual data like photos and videos are thought of as a representation in the spatial domain, we can then apply transformations to represent them in another domain. By decomposing an image into low and high frequencies, distinct features emerge that may have otherwise been hidden. Low frequencies capture the foundational elements of an image, such as colors, highlights, and tones, while high frequencies encode textural details, including edges and contours. This principle has been widely leveraged in photography and image editing software like Adobe Photoshop, which uses such techniques to apply effects like sharpening, blurring, and other retouching adjustments [4].

In signal processing, transformations like the Fast Fourier Transform (FFT) have long been used to convert time-domain signals, such as those from oscilloscopes, into the frequency domain for spectral analysis. Similarly, the Discrete Cosine Transform (DCT) has proven effective in compression algorithms.

A particularly promising approach is the Walsh-Hadamard Transform (WHT), a simple yet powerful mathematical technique. The WHT is based on a 2×2 matrix that can be recursively multiplied to form larger matrices. This structure imparts several desirable properties for machine learning and computational applications, including orthogonality, symmetry, and scalability. Its computational complexity often scales with the square root of the data length, making it highly efficient.

The WHT has a long history of practical application in areas of dimensionality reduction. During the 1960s and early 1970s, NASA used the Hadamard transform as a basis for compressing photographs transmitted from interplanetary probes [5]. More recently, the WHT has been successfully applied across diverse domains, including evolutionary search algorithms, gradient calculations, and regularization techniques, often outperforming traditional linear methods in both efficiency and effectiveness [6].

## II. Research Topic

The motivation for this project stems from the need to develop efficient methods for reducing the computational burden of deep learning models. In this study, we aim to implement the Walsh-Hadamard Transform (WHT) and explore its integration into an image classification Convolutional Neural Network based on the ResNet-50 architecture. Specifically, we will evaluate the performance of the WHT at different layer positions and compare its performance to two other transforms identically placed in the model: the Fast Fourier Transform and the Discrete Cosine Transform.

Our hypothesis is that these transforms can extract highly representative image features that may not be easily interpretable by conventional neural networks typically used in image processing tasks. By leveraging these transforms, we aim to reduce computational demands while maintaining model performance.

## III. Previous Work

There have been several papers written in the domain of transform based reductions in neural networks. One such paper that we took inspiration from was "Deep Neural Network With Walsh-Hadamard Transform Layer for Ember Detection" [7]. The goal of the paper was to develop a neural network for the purpose of detecting embers in an IR video for the sake of wildfire prevention. The researchers accomplished this by implementing a model using the Walsh-Hadamard Transformation (WHT) and another model with the Fast Fourier Transform (FFT). The transforms were placed at the input of their model, followed by ResNet-18 and ResNet-34. The study found that WHT performed better than an implementation with FFT and that ResNet-18 was sufficient for the task [7].

Another previous study that we analyzed during our initial research was on proving the efficacy of the Hadamard Transform in binary neural networks [8]. The researchers compared WHT with Discrete Cosine Transform (DCT) aiming to optimize the floating-point arithmetic that was commonly used in the input layer, finding that both transformations were functionally equivalent but had alternative mathematics. By utilizing ResNet-18 and a custom MobileNet-based neural network inspired by the ReActNet architecture, the study observed a slight decrease in accuracy compared to the baseline model [8]. However, the WHT outperformed the DCT by significantly reducing the computational effort required to train and fit the models [8].

Lastly, a third study we reviewed, was conducted by the same group of researchers—H. Pan, D. Badawi, and A. E. Cetin—investigated replacing certain convolutional layers in deep neural networks with binary-based WHT layers [9]. The study utilized both one-dimensional and two-dimensional WHTs to replace the convolutional operations. To evaluate the efficacy of the WHT layers, the researchers tested four neural networks: MobileNetV2, MobileNetV3, ResNet-20, and ResNet-34 [9]. The results showed that, for most of the tested networks, the WHT layers led to a reduction of at least half of the trainable parameters, with a corresponding decrease in accuracy [9].

In this study, we aim to explore the use of FFT, DCT, and WHT, building on methodologies from previous research. However, unlike prior studies, we will integrate these transformations into a more complex model—ResNet50—as the base architecture. While past research has typically been limited to exploring only two transformations, our approach includes all three. Furthermore, we are investigating the implementation of these transforms at three distinct locations within the model: the input layer, early layers, and a combination of early and later layers. By focusing on key metrics such as computational efficiency, power consumption, and classification accuracy, this study examines the impact of these transformations in novel configurations. The diverse combinations of transformations and their placement within the model enable us to present a unique and comprehensive analysis.

## IV. TRANSFORMATIONS

### A. FFT

The Fast Fourier Transform (FFT) is an approximation algorithm for the Discrete Fourier Transform (DFT), first introduced by W. Cooley and John Tukey in their 1965 paper "An algorithm for the machine calculation of complex Fourier series". Fourier transforms allow the transformation of signals from one domain (such as time or space) into another domain (frequency), and are foundational to modern compression algorithms in both audio and imagery. The Fourier analysis originates with the proposal of Jean-Baptiste Fourier, who in 1807 claimed that any continuous periodic signal could be represented by the sum of properly chosen sinusoidal waves [10].

The Fast Fourier Transform employs a recursive method based on matrix multiplication, enabling it to achieve a computational complexity of O(NlogN) compared to the original DFT's O(n$^2$) [11]. This significant improvement in efficiency has made FFT monumental in approximating derivatives, noise reduction, communication compression, and a spectrum of tasks in Fourier analysis.

The FFT algorithm is based on the Fourier exponential series, defined by the equation:

The Discrete Fourier Transform (DFT) of a vector **X** is defined as:

$$X[k] = \sum_{n=0}^{N-1} x[n] \cdot e^{-i\frac{2\pi}{N}kn}, \quad k = 0, 1, \ldots, N-1.$$

This can be represented in matrix form as:
$$\mathbf{X} = \mathbf{F} \cdot \mathbf{x},$$
where **F** is the Fourier matrix:
$$\mathbf{F}_{m,n} = e^{-i\frac{2\pi}{N}mn}. \tag{1}$$

When written as a function, it becomes the Fourier Transform, which takes a real valued input (typically representing samples integrated over time) and converts it into a frequency domain representation. This frequency domain output has two components: magnitude and phase. The magnitude indicates the strength of a frequency, while the phase provides the offset of that frequency. An inverse operation, called the Inverse Discrete Fourier Transform (IDFT), can reconstruct the original signal from its frequency components [12].

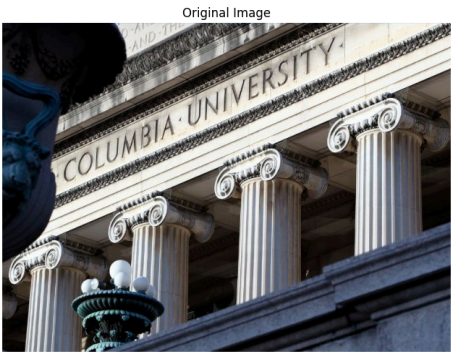

Figure 1: Original RGB Image

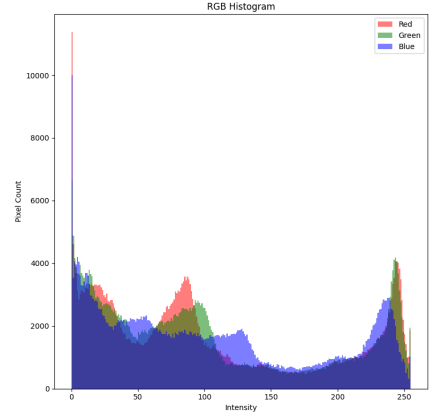

Figure 2: RGB Image Displayed Chanel Wise as a Histogram

#### 1) FFT in Images

Although images are not inherently in the time domain, they can be thought of as existing in the spatial domain, where pixel intensities represent spatial information. The Fourier Transform can be applied to convert this spatial information into the frequency domain, expressed as sums of sine and cosine waves at various frequencies.

Consider an example of an image used for ResNet50 - a 232x232 pixel RGB image. In vector form, this image is represented as (232, 232, 3), where the dimensions correspond to height, width, and color channels. For this example, we will simplify this by converting to grayscale, reducing it to a single channel (232,232,1). A 2D Fourier Transform is then applied, resulting in two outputs: the magnitude and the phase.

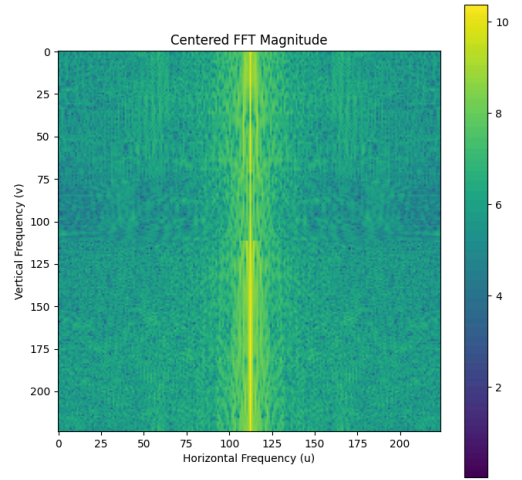

Figure 3: Magnitude: Represents the intensity of various frequencies in the image. This component highlights the broad structures and smooth areas.

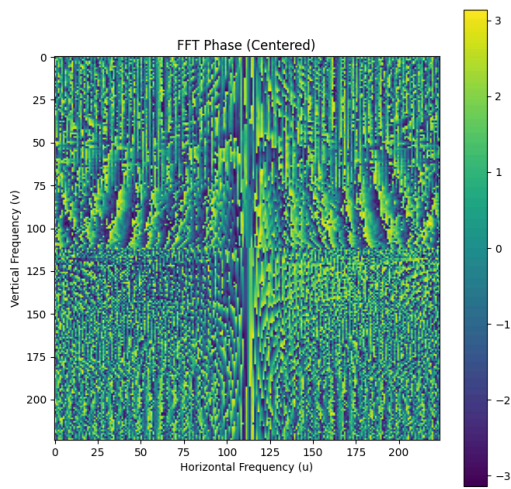

Figure 4: Phase: Encodes the positional information of those frequencies, critical for reconstructing the image's spatial details.

The Phase representation is particularly important in signal analysis, as it preserves the relative positioning of features within the data. For instance, in image reconstruction or recognition, phase information can maintain structural integrity even if the magnitude is degraded.

With the magnitude and phase split out, we can manipulate the frequency components to achieve specific effects - effectively a band pass filter. By masking out higher frequencies to isolate the low-frequency components, which correspond to smooth regions such as tones, gradients, and large-scale color changes. Masking out lower frequencies isolates the high-frequency components, enhancing details such as edges, textures, and fine patterns. This is commonly used in noise reduction, where selectively masking high-frequency noise preserves the essential low-frequency information.

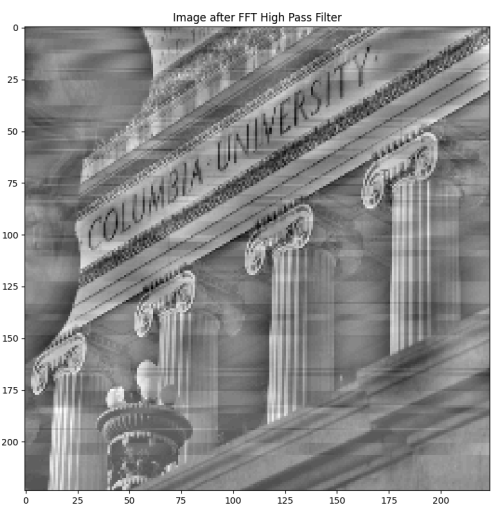

Figure 5: Image after FFT High Pass Filter

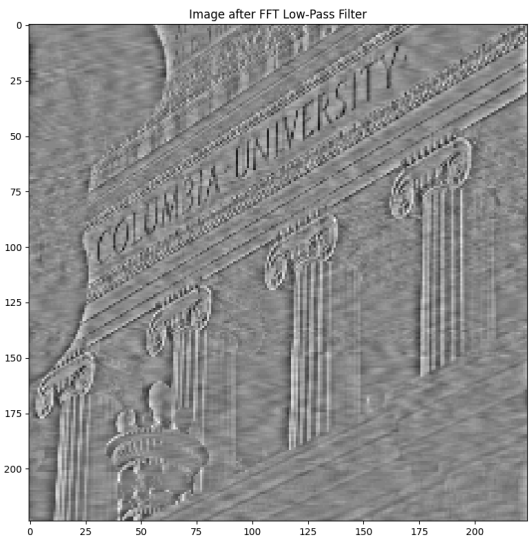

Figure 6: Image after FFT Low Pass Filter

It can be difficult to image an image in the frequency domain, so an analogy for frequency filtering is a bas-relief sculpture: the low-frequency components represent the overall shape and form, while the high-frequency components define the fine details and texture.

*B. DCT*

The Discrete Cosine Transform (DCT) is a linear and orthogonal transformation that represents a finite sequence of data points as a weighted sum of cosine basis functions with varying frequencies [13]. It is widely used in image and video compression due to its ability to compactly represent signal information, preserving critical data while discarding redundancy. Early video coding standards, such as MPEG-1, MPEG-2, and H.263, leveraged DCT for intra-frame compression [14]. By transforming data into the frequency domain, DCT is able to take spatial information and split it into frequency components, making it easier to identify less human perceptual details.

Previous studies introduced the Block Walsh-Hadamard Transform (BWHT), which uses a methodology similar to the DCT employed in JPEG compression. In this method, the input is divided into smaller blocks, and the Walsh-Hadamard Transform is applied to each block individually. This block-wise approach minimizes the need for zero-padding, as the transformation is applied to each discrete section independently, enhancing computational efficiency.

The DCT was first introduced in 1972 by Nasir Ahmed, who proposed it to the National Science Foundation as a study of the cosine transform using Chebyshev polynomials [15]. Since then, it has become a cornerstone in image processing and compression technologies, enabling significant advancements in digital media formats. Among the various DCT formulations, the DCT-II variant is particularly prominent and serves as the focus of this study.

Normally, direct computation of the DCT involves $O(N^2)$ complexity due to the nested summations. However, TensorFlow optimizes this process by leveraging the Fast Fourier Transform (FFT), achieving a reduced complexity of $O(N\log N)$. This optimization arises from the mathematical relationship between DCT and the Discrete Fourier Transform (DFT). By applying a DFT to an even-symmetric version of the input data, the results can be adapted to match the outcomes of the DCT, significantly improving computational efficiency.

If we have a sequence of $x_0, x_1, x_2 \ldots x_{n-1}$ using DCT-II we will get the following sequence as our output $X_0, X_1, X_2 \ldots X_{N-1}$ where [16,17,18,19]

$$X_k = \sum_{n=0}^{N-1} x_n \cos[\frac{\pi}{N}(n + \frac{1}{2}k], \; k = 0,1,\ldots, N-1 \quad (2)$$

when there is no normalization factor α(k) [16]. DCT-II of the input vector x[n] can be calculated as:

$$X[k] = \sum_{n=0}^{N-1} x[n]\cos(\frac{\pi}{N}(n+\frac{1}{2})k), \; k = 0, 1, \ldots, N-1 \quad (3)$$

Since we are using the norm = "ortho" in our codes, a normalization factor α(k) is applied, making this transform orthonormal. After the normalization, our output changes into:

$$X[0] = \sqrt{\frac{1}{N}} \sum_{n=0}^{N-1} x[n] \cos(\frac{\pi}{N}(n+\frac{1}{2}) \cdot 0) = \sqrt{\frac{1}{N}} \sum_{n=0}^{N-1} x[n] \quad (4)$$

For the values of k that are greater than 0 it can be said that:

$$X[k] = \sqrt{\frac{2}{N}} \sum_{n=0}^{N-1} x[n] \cos(\frac{\pi}{N}(n+\frac{1}{2}) k) \; [5] \quad (5)$$

After applying a DFT to an even version of the original dataset the results are mathematically similar after DCT implementation. A point $x_n$ can be observed after the DFT implementation as:

$$X_k^{DFT} = \sum_{n=0}^{N-1} x_n e^{-j\frac{2\pi}{N}nk} \quad (6)$$

Because after DFT implementation only cosine values are left we can create a 2N-point sequence of $y_n = \{x_n, 0 <= n < N; \; x_{2N-1-n}, \; N <= n < 2N\}$ (6) Tensorflow implementation utilizes the equation (6) and constructs a 2N-point sequence based on FFT. If we have an input x[n], the implemented 2N-point sequence y[n] can be observed as:

y[n] = x[n], n = 0, …, N-1; y[2N-1-n] = x[n] where n = 0,…, N-1  (7)

Which as we mentioned earlier created an even version of the original dataset. After that, we can compute the FFT of y[n] as [19]:

$$Y[K] = \sum_{n=0}^{2N-1} y[N] e^{-j\frac{\pi}{N}nk} \text{ where } k = 0, \ldots, 2n-1 \quad (8)$$

By utilizing the Euler's formula $\cos(\theta) = (e^{j\theta} + e^{-j\theta})$ we can obtain the following:

$$X[K] = \text{Re}\{e^{-j\frac{\pi}{2N}k} Y[k]\} \text{ where } k = 0, \ldots, N-1 \quad (9)$$

The coefficients after FFT transformation are selected by the equation above.

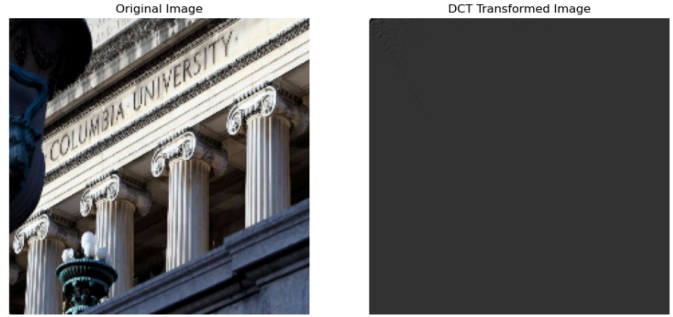

Figure 7. Side-by-side comparison of Original Image and DCT Transformed Image

The original image on the left side has high details and sharpness, whereas the pixels in the output on the right side don't correspond to the pixel brightness as the spatial domain but store information about the frequencies.

As expected, the DCT-transformed image is dark and we can't see visually the information that it stores. In the frequency domain, the change in the pixel values is referred to as frequency.

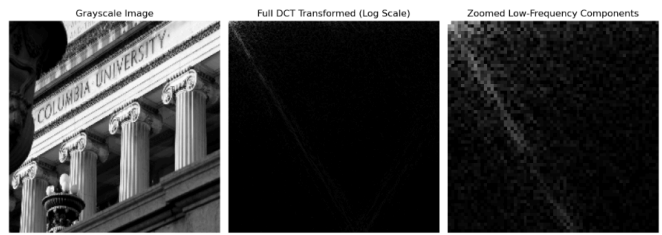

Figure 8. Side-by-side comparison of Grayscale Image, Full DCT Transformed Image on log scale, and zoomed low-frequency components

The output of our custom function can be observed in Figure 8. We have generated three images, respectively: the grayscale image of the original image, the DCT Transformed Image on a logarithmic scale, and the zoomed low-frequency components. We transformed the original image into the grayscale format to solely focus on the brightness rather than the color channels, as the DCT implementations only operate in the frequency domain and do not utilize color channels. As can be seen from the middle image, Full DCT Transformed in Log Scale, regions with the most energy are concentrated in the top-left corner. This is expected because low-frequency layers tend to dominate over high-frequency components. In the last image, we can observe the low-frequency region more clearly. We achieved this by cropping the DCT matrix to a size of 64x64, focusing on the top-left components. This part is the most essential in our image because it represents the most significant features. We have tried using different images to observe this phenomenon, as you will see in Figure 9.

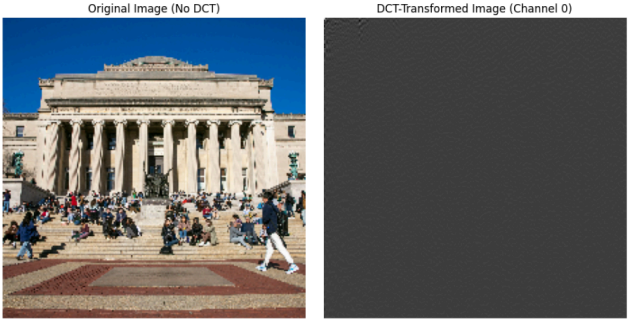

Figure 9. Columbia University original image and DCT-Transformed Image side-by-side

In image processing, DCT converts spatial-domain images into the frequency domain. While DCT-transformed images may appear less interpretable to the human eye, they concentrate critical information in low-frequency components. This property allows for efficient compression, as high-frequency components, which often correspond to noise or minor details, can be discarded without significantly affecting visual quality.

## C. WHT

The Walsh-Hadamard Transform (WHT) is a class of Fourier transforms applied to data, combining elements from the works of Jacques Hadamard and Joseph L. Walsh. The Hadamard matrix, first introduced in 1893 by Jacques Hadamard in his paper "Résolution d'une question relative aux déterminants", serves as the foundation for the recursive structure of the WHT [21]. Similarly, Walsh functions, developed and published by Joseph L. Walsh in 1923, contributed to the alternating patterns used in the transform [22]. Together, Walsh functions and the Hadamard matrix form the basis of the WHT.

Nasir Ahmed, who proposed the DCT, explored the use of the Walsh-Hadamard Transform for 6:1 compression in interframe transform coding for NASA Ames Research Center [23]. This implementation demonstrated the WHT's early potential in data compression.

The WHT recursively applies the Hadamard matrix to a dataset as its size increases. Constructing a Hadamard matrix involves using Sylvester's Construction [24], a recursive method where smaller matrices are used as components of larger matrices.

$$H_n = \begin{bmatrix} H_{n/2} & H_{n/2} \\ H_{n/2} & -H_{n/2} \end{bmatrix} \quad [8]\ (10)$$

Where $H_1 = 1$ [8] (11)

This method is particularly efficient for datasets with dimensions that are powers of 2 ($2^n$), as the matrix size doubles with each iteration. However, it also requires datasets that are not that size to be resized.

$$\text{For } X \in R^W, W = 2^N \quad (12)$$

$$\text{For } X \in R^{H \times W}, H \times W = 2^N \times 2^N \quad (13)$$

Once the Hadamard matrix is formed, it must be normalized before being applied to data. Normalization is achieved by dividing each element of the matrix by the square root of its size or the length of one side of the matrix, as shown here:

$$H_W = \frac{1}{\sqrt{n}} H_n \quad [7]\ (14)$$

This ensures that the transform maintains numerical stability and consistency across different applications.

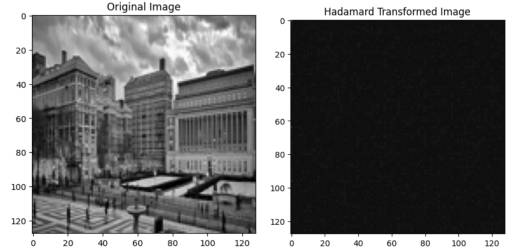

Figure 10. Original Image and Hadamard Applied Image

After applying the walsh-hadamard matrix and inverse walsh-hadamard matrix to the same image we aim to get the image without any corruption. As can be seen in Figure below, it can be seen that our Walsh-Hadamard function works as intended.

In one dimension data, the Hadamard matrix can be directly multiplied by the dataset. For image processing where the data represents two dimensional spatial information, the matrix is applied separately to both rows and columns and subsequently for each RGB channel. The WHT's computational simplicity stems from its reliance on matrix multiplication. This efficiency makes the WHT particularly advantageous for matrix multiplications in GPU based processing like deep learning and signal processing.

## D. Transformations in CNNs

Convolutional Neural Networks (CNNs), widely used in image processing, extract meaningful features by applying convolutional filters across the spatial dimensions of an image. Integrating transformations such as WHT, DCT, and FFT into CNN workflows offers the potential to enhance the network's ability to analyze images by leveraging frequency-domain information alongside spatial data.

Applying a transformation like DCT, WHT, or FFT at the input stage can provide additional frequency-domain insights, complementing the spatial features captured by the CNN. This preprocessing step may help the network better represent and interpret the underlying structure of the data.

Introducing transformations at intermediate layers can simplify complex patterns into distinct frequency components, potentially accelerating convergence during training. By reducing redundancy and emphasizing critical features, these transformations may improve the efficiency of feature extraction and learning.

Incorporating a transform at the output layer can refine the activation layers by leveraging frequency-domain insights to enhance prediction accuracy. This approach could help the network make more nuanced decisions by capturing subtle patterns and relationships in the frequency domain that are not apparent in the spatial domain alone.

## V. Model Architecture

Hyperparameters such as layer type, placement, and depth have a significant impact on a neural network model's behavior and performance. To explore these effects, we implemented three different models with varying layer placements for each of the transformations.

### A. ResNet Architecture

Popularized in 2015 by Microsoft researchers through the paper Deep Residual Learning for Image Recognition [25], the ResNet model was selected for its proven performance, ease of use as a pre-built model, and availability in a pre-trained format. ResNet-50, a variant of the deeper 152-layer ResNet, is a 50-layer convolutional model with 23.6 million parameters designed for image classification tasks with a maximum input size of 224 × 224. The variant we are using is readily available through TensorFlow, allowing seamless integration into various model architectures. Moreover, the model comes pre-trained on the expansive ImageNet dataset.

Pre-trained models enable a technique called transfer learning, where a model is initially trained on one dataset (e.g., ImageNet) and then fine-tuned for a different dataset. The model's weights—parameters that adjust to the learned features, are adjusted during training to align with the underlying distribution of the new data. However, these weights can also be "frozen," meaning they remain fixed and are not updated during training. Freezing weights is critical because training on large datasets like ImageNet is computationally expensive and time-consuming, and transfer learning allows leveraging this prior training.

In this study's implementation, we use ResNet-50 as a base model for transfer learning. The process begins by training the model on the new dataset with the pre-trained weights frozen, allowing only the shallow or later layers to update and adapt to the new data. This step ensures that the deeper, more complex layers retain the robust features learned from the original training. Once this initial training stabilizes, the weights are "unfrozen," and the entire model is trained again. This two-phase approach ensures that the model benefits from both the pre-trained features and the specificity of the new dataset, avoiding significant alterations to the pre-trained weights that could negate the advantages of transfer learning.

### B. Base Model

Similar studies [26, 27] utilized ResNet-50 as their primary base model, and this study adopted it as a point of comparison. Initially, the MNIST and CIFAR-10 datasets were used due to the widespread availability and usage in machine learning studies. However, since these datasets were rescaled from 32x32 to 224x224 for model compatibility, the complexity of the ResNet-50 model caused it to easily overfit by the fifth epoch.

To enhance data generalization, multiple techniques were explored, including L2 regularization, data augmentation, freezing and unfreezing layers, and transfer learning. However, none of these methods sufficiently mitigated overfitting, making evaluations challenging.

Given the model's large learning capacity, CIFAR-100 was finally chosen as the primary dataset for evaluation. Benchmarking results from online sources [28] were used for comparison and insight.

The initial ResNet-50 implementation utilized a pre-trained ImageNet model without applying data augmentation, dropout layers, or regularization. This design provided initial insights into the model's performance with CIFAR-100. As shown in Figure 14, the model exhibited heavy overfitting after the 13th epoch, with training accuracy continuing to rise while validation accuracy plateaued. Despite the overfitting, the model achieved a training accuracy of 0.6618 with a loss of 1.5186, which closely matched the benchmark results [28]. The test accuracy was similarly consistent, at 0.6608 with a test loss of 1.5223.

In addition to evaluating the base model on RGB data, we recognized that the transforms applied in our experiments, such as Fast Cosine Transform (FCT) and Discrete Cosine Transform (DCT), resulted in entirely different data representations. These changes made the standard approach of freezing early layers during transfer learning ineffective. To address this, all layers were set as trainable from the start, allowing the model to adapt and converge effectively under the new conditions.

The final base model used for transformation experiments permitted full training of parameters. It incorporated 2D average pooling followed by a substantial 0.5 dropout layer at the output of the ResNet-50 model. The final output layer consisted of a fully connected dense layer with 100 neurons, softmax activation, and an L2 kernel regularizer of 1e-4.

The model was compiled using the Adam optimizer and sparse categorical cross-entropy as the loss function.

### C. Model 1

In this implementation, the transformation layer was applied directly after the input layer, serving as an additional feature extraction mechanism during the data preprocessing stage. This approach, shown in Figure 12, aimed to enhance feature extraction early in the pipeline.

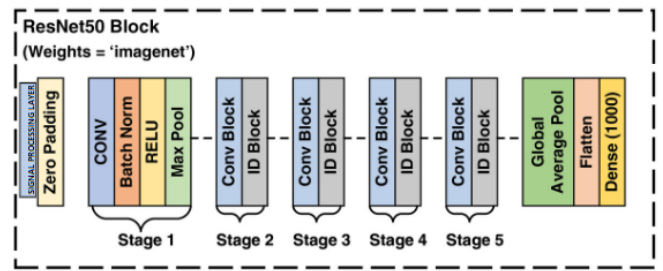

Figure 11: First design choice of ResNet-50 model with Signal Processing algorithms implemented

### D. Model 2

The second implementation integrated the transformation layer into the early (conv2_block1) convolutional blocks of the ResNet-50 architecture. Early convolutional blocks in ResNet-50 typically detect simple features like edges and textures. Placing the transformation here allows these blocks to operate in the frequency domain in the case of FFT and DCT, potentially enabling the model to learn more meaningful, otherwise hidden patterns [29]. Additionally, early convolutional layers help reduce image noise, improving their performance on noisier inputs [30]. By integrating these transformations into the early layers, the model reduces computational complexity by dividing feature maps into smaller groups [31].

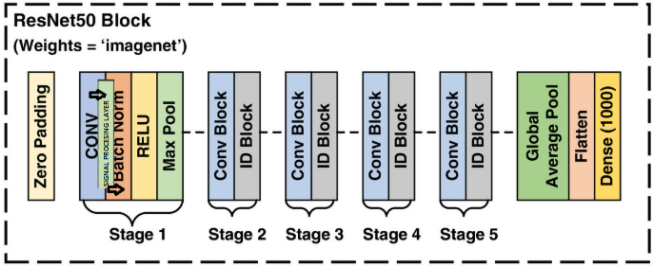

Figure 12: Second design choice of ResNet-50 model with Signal Processing algorithms implemented

### E. Model 3

The third implementation combines aspects of both earlier models by employing two transformation layers: one in the early layers (as in Model #2) and another after the later convolutional blocks (conv4_block6). This hybrid approach leverages the benefits of both preprocessing transformation and trainable transformation layers within the ResNet-50 architecture. While previous studies [26, 27] have explored multi-layer utilization of transform layers, our model is novel as it combines a ResNet-50 backbone with multi-layer transformations. This dual-layer implementation aims to capture both low-level and high-level features.

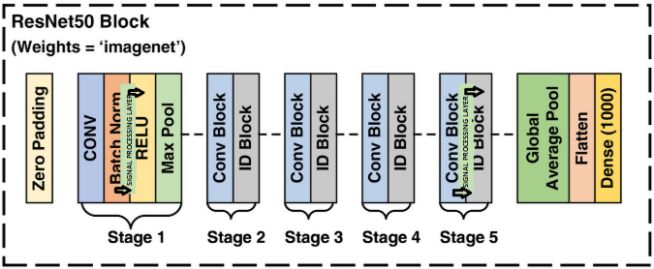

Figure 13: Third design choice of ResNet-50 model with Signal Processing algorithms implemented:

## VI. TRANSFORM IMPLEMENTATION

### A. FFT

The FFT layer was implemented using TensorFlow's Keras Layer class, utilizing the built-in fft2d function to compute the 2D Fast Fourier Transform. This function operates on the last two axes of the input tensor, perfect for image use, and uses an optimized fast algorithm to perform the transformation efficiently.

The class expects the layer's input shape to be defined as [B, H, W, C], where B is the batch size, H and W are spatial dimensions, and C is the number of channels. In the call method, the input tensor is cast to a required complex data type before applying fft2d. The magnitude of the resulting complex output is then computed and returned as a float32 tensor as required by the model.

---
**Algorithm 1** Fast Fourier Transform 2D Layer
**Require:** Tensor $T$ of shape (batch, height, width, channel)
**Ensure:** Transformed tensor
  Initialize:
    Cast input $T$ to `complex64` datatype
  Apply 2D FFT using TensorFlow:
    $T \leftarrow$ TensorFlow 2D FFT$(T)$
  Take the absolute value of the result:
    $T \leftarrow |T|$
  Cast result back to `float32` datatype:
    $T \leftarrow \text{Cast}(T, \texttt{float32})$
  **return** $T$
---

To verify correctness, the layer's functionality was tested. A test tensor was passed through the FFT layer implementation, and its output was compared against the expected result, directly calculated using TensorFlow's fft2d and math.abs functions. This confirmed the accuracy of the class implementation for the frequency domain transformation.

### B. DCT

A custom Discrete Cosine Transform (DCT) function was implemented using TensorFlow's tf.signal.dct library. However, TensorFlow's implementation only supports 1D DCTs along a specified axis. Since 2D spatial data is required for image processing, additional steps were necessary to extend the functionality.

To perform a 2D DCT, the tensor was first transposed from its original shape [batch, height, width, channel] to [batch, height, channel, width] to bring the width dimension to the last axis. The 1D DCT was then applied along this axis, followed by transposing the tensor back to its original shape. Similarly, the 1D DCT was applied along the height axis by transposing the tensor to [batch, width, channel, height], applying the transform, and transposing it back to its original shape.

---
**Algorithm 2** Discrete Cosine Transformation 2D Layer
**Require:** Tensor $T$ of shape (batch, height, width, channel)
**Ensure:** Transformed tensor
  Initialize the layer:
    Call the parent function
  Define the Forward Pass:
    Input tensor shape [batch, height, width, channels]
  Perform DCT along the width axis:
    Transpose the input tensor to [batch, height, channels, width]
    Apply the 1D DCT transformation along the width axis
    Transpose back to [batch, height, width, channels]
  Perform DCT along the height axis:
    Transpose the input tensor to [batch, width, channels, height]
    Apply the 1D DCT transformation along the height axis
    Transpose back to [batch, height, width, channels]
  **return** Transformed tensor
---

To evaluate the correctness of this implementation, Parseval's Theorem [32] was used. This theorem establishes a relationship between the energy in the spatial domain and the frequency domain, stating that the integral of the square of a function in the spatial domain equals the integral of the square of its Fourier coefficients in the frequency domain.

The function was tested on an image of size 763×556 pixels, yielding a frequency domain energy of 39923.2656 and a spatial domain energy of 39923.2617. The minimal difference of 0.004 is attributed to numerical rounding in Python's implementation. These results confirm that the custom 2D Discrete Cosine Transform implementation works as intended, maintaining energy equivalence between the spatial and frequency domains.

*C. WHT*

A custom Walsh-Hadamard Transform (WHT) function was implemented, as it is not readily available in existing libraries, unlike FFT and DCT. This function was developed using TensorFlow's Layer class and consists of three primary components: one to construct the recursive Hadamard matrix, another to normalize it, and a third to handle data reshaping and transformations.

The implementation begins by reshaping the input tensor, originally shaped as [batch, height, width, channel], into [batch, channel, height, width] through a transposition operation. The batch and channel dimensions are then flattened into a single combined dimension, resulting in a 3D tensor, where this combined dimension represents the product of the batch size and the number of channels.

---

**Algorithm 3** Walsh-Hadamard Transformation Layer
---
**Require:** Tensor $T$ of shape (batch, height, width, channel)
**Ensure:** Transformed tensor
   Initialize:
      Set size = size
      Generate normalized Hadamard matrix of size = size
      Store it as a non-trainable TensorFlow variable
      Convert the input tensor to the same datatype as normalized Hadamard matrix
   Get the input tensor dimensions
   Define the Forward Pass:
      Transpose inputs to [batch, channels, height, width]
      Convert channels and batch into a single dimension
      Apply the Walsh-Hadamard transform along the width axis
      Transpose the output to work with height dimension
      Transpose back to the original form
   **return** Transformed tensor

---

Next, the Hadamard matrix is applied iteratively along the height dimension, treating the input tensor as blocks of two-dimensional matrices. Once this operation is complete, the tensor is transposed to reorder its dimensions, enabling the application of the Hadamard matrix along the width dimension.

Following these transformations, the tensor is returned to its original shape by reversing the transpose and reshape operations. The final output is reshaped back into its initial four-dimensional format, [batch, height, width, channel], ensuring compatibility when integrated into models as an inner layer, particularly for models two and three.

To validate the correctness of this implementation, tests were conducted on a 2D representative input. The transformed data was inverted using the inverse Walsh-Hadamard Transform, and the absolute values of each element were compared to the original input. The results showed agreement, confirming that the custom Walsh-Hadamard Transform function was accurately constructed across both spatial dimensions.

## VII. DATASET

This study selected the CIFAR-100 dataset for our experiments due to its accessibility and diversity in classification tasks. With 100 classes and over 60,000 images, the dataset provides a comprehensive benchmark for evaluating the performance of our approach. However, the relatively low resolution of CIFAR-100 (32x32 pixels) is a limitation, as a higher resolution closer to the ResNet input size of 224x224 would have been more suitable for testing larger network architectures like ResNet.

An additional advantage of using the CIFAR-100 dataset is its seamless integration into TensorFlow, allowing it to be easily accessed through its keras.datasets module. This streamlined integration facilitates efficient dataset loading and preprocessing, and most importantly reproducibility of the study.

## VIII. METRICS

The metrics for this study were clearly defined based on its objectives: investigating computational efficiency, power usage, and accuracy between the three transformations. These key metrics include classification accuracy, training power consumption, training time, and inference time. While some loss of accuracy is anticipated due to dimensionality reduction, the study focuses on assessing the feasibility and effectiveness of these transformations as methods for reducing the computational demands in neural networks. By systematically evaluating these performance metrics, we aim to determine whether such transformations are useful for improving efficiency in deep learning applications.

Although classification accuracy is a standard metric available in most machine learning libraries, GPU utilization, memory usage, and power consumption are not, To address this, a monitoring function was developed using NVIDIA's NVML library which allows parameters to be directly read such as power consumption, GPU memory usage. This function logs and visualizes these additional metrics and integrates seamlessly into the TensorFlow callback system.

In addition to the study-specific metrics, training accuracy and loss were also analyzed to gain insight into the training process, identify trends, and observe potential overfitting in the model.

## IX. ENVIRONMENT

This study was implemented in Python 3 using Jupyter Notebook and Google Colab. TensorFlow, with its Keras API, was chosen as the primary machine learning framework. Given the dataset size and the complexity of the 50-layer ResNet50 model, we utilized the NVIDIA A100 GPU platform. Lower-tier GPUs, such as the NVIDIA L4 and T4 which typically have 16GB of GPU memory, encountered memory overflow issues during training, as an unrestricted run typically required approximately 25 GB of GPU memory. The NVIDIA A100 platform, with its 40 GB memory in the standard configuration available on Colab and Google Compute Platform, was thus necessary to handle the model's demands effectively.

Final tests were conducted on a Colab instance running Python 3.10, TensorFlow 2.17, and Keras 3.5 utilizing an NVIDIA A100. Training ranged from 10 to 90 minutes, depending on the specific model and configuration.

## X. RESULTS

In this chapter, we will compare the results with and without implementing these three signal-processing techniques with the evaluation metrics we have discussed earlier. After training the first ResNet50 base model which also can be seen in Figure 14. We discussed the results after the signal processing implementations.

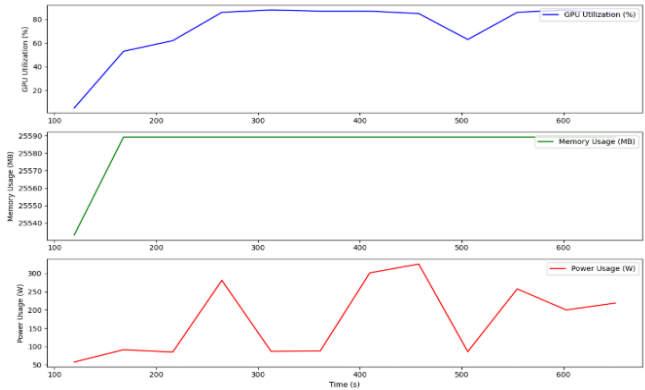

Figure 14. Baseline Model ResNet50 power consumption

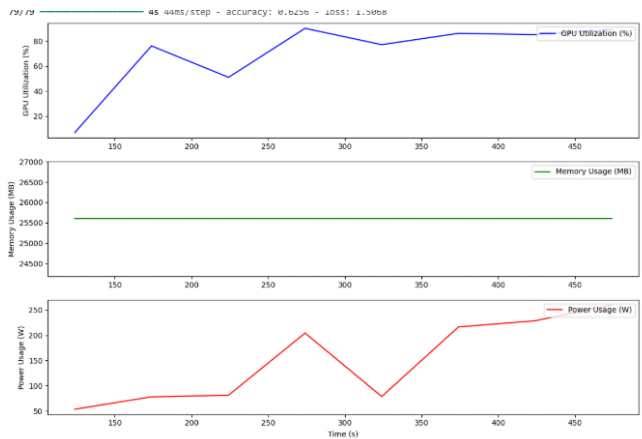

Figure 15. First Implementation of 2D DCT Layer

In these two models, we used the same hyperparameters. The model with a 2D DCT layer implemented shows a more stable power consumption profile compared to the model without utilizing this transformation. We can see that both models show similar memory usage, which implies that using a 2D DCT layer does not impact memory consumption. When we compare the accuracies of these different models, we observe that the validation accuracy for the base model was 0.6618, while the validation accuracy of the 2D DCT model was 0.6256. As shown in the previous papers [25,26], utilizing signal processing techniques with a slight accuracy loss decreased the GPU consumption. By placing the first DCT Layer after the early convolutional blocks and the second DCT Layer in the intermediate stages we observed the following results: The accuracy, which was 0.6618 in the baseline model and 0.6256 in the first 2D DCT implementation, dropped significantly to 0.3531. We believe that the complexity and resolution of CIFAR-100 made frequency-domain features less dominant, resulting in the observed drop in accuracy.

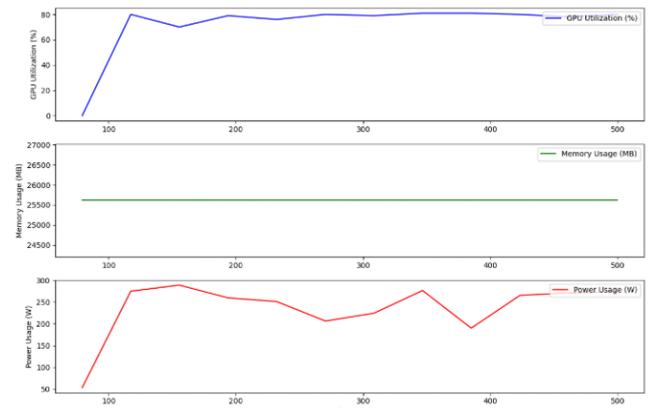

Figure 16. Placing DCT in early and later stages

Although the accuracy did not improve, we managed to maintain the memory consumption at a stable level, providing additional insight that using discrete cosine transformations does not consume additional memory. Following this implementation, we proceeded to evaluate the third implementation of the 2D DCT layer by placing it after an early convolutional block without changing the hyperparameters. In this configuration, we observed that the accuracy increased to 0.5977, with a testing loss of 1.5615.

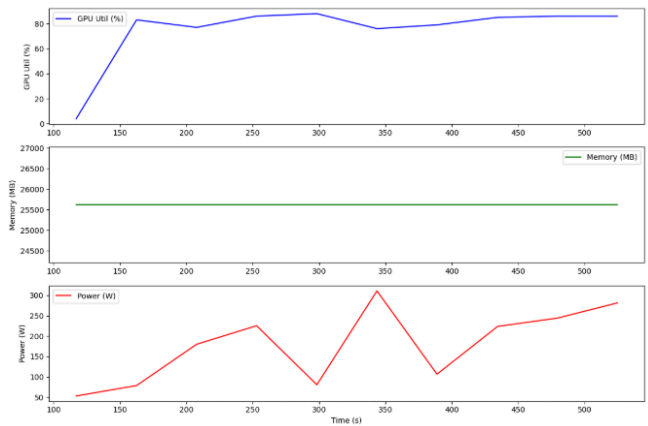

Figure 17. Third Implementation of 2D DCT Layer

As we look at Figures 8, 9, 10, and 11, we can see that the memory consumption remains stable while training with DCT layers. This indicates that the 2D DCT layer does not consume any additional memory. However, in certain DCT placements, we observe a decrease in power consumption. During this project, we realized that adding a 2D DCT layer acts as an additional hyperparameter, which can reduce power consumption when applied to the right place in the right dataset.

After the first FFT implementation, we observed that the model's accuracy remained well below 10%. Although it initially decreased from extremely high values, it stabilized at relatively large levels. There was also fluctuation in validation accuracy throughout the testing phase (ranging between 0.03 and 0.0052), showing no clear trend that indicates effective learning. GPU utilization, on the other hand, started at a low value of 6% but stabilized at 80% usage by the end of the training. Unlike DCT and WHT, we observed some changes in memory usage, which hovered around 5001 MB during training. GPU power usage showed occasional spikes, reaching up to 269.99W. In total, our integrated GPU usage was calculated as 124.20 GPU-seconds, and the total integrated power consumption was 29,278.94 Joules.

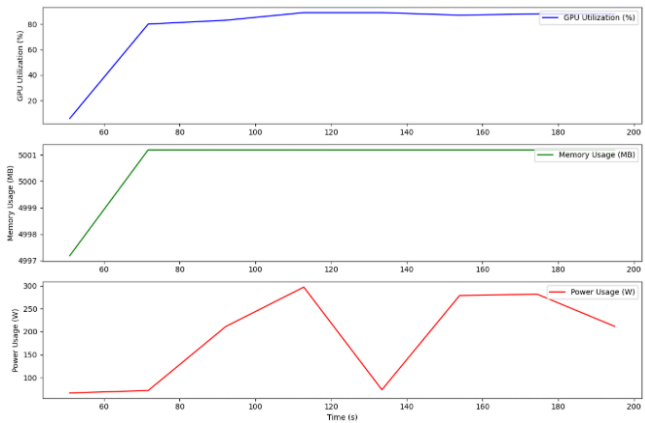

Figure 18. First Implementation of FFT Layer

After unlocking the weights in our base ResNet-50 architecture, we concluded the training with the same hyperparameters. The model showed an increase in training accuracy after 15 epochs, reaching 73.9%, with a final training loss. However, the fluctuations in validation accuracy remained unchanged, indicating that the model was still overfitting. Similarly, GPU utilization started at 6% and stabilized between 78–81% by the end of the training. Memory usage remained consistent at around 15,301.19 MB. Power consumption exhibited some spikes during training, peaking at 270.49 W, with total power consumption recorded at 121,684.85 Joules.

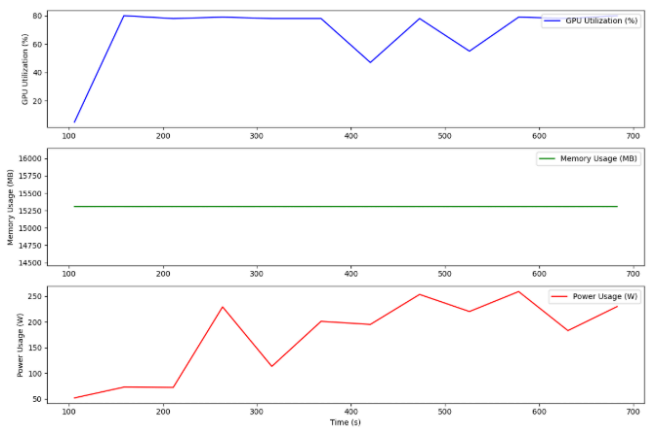

Figure 19. Second Variant, First Implementation of FFT Layer

After implementing the second FFT model, the accuracy increased from %15.5 to over %83.3 by Epoch 12, validation accuracy on the other hand, improved significantly in early training stages and then peaked at around %59.97 making this version of the model with highest overall performance. GPU memory usage stayed consistently around 15 GB, like the previous model GPU utilization hovered around %78 to %80 for most epochs. Total integrated GPU usage was 424.82 GPU-seconds and the total integrated power consumption was 106,346 Joules.

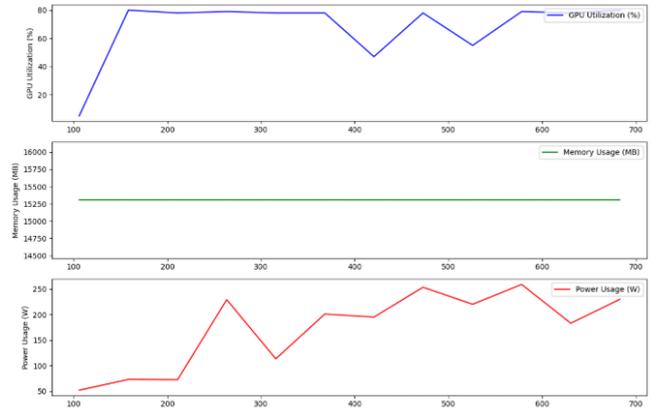

Figure 20. Second Implementation of FFT Layer

In our final FFT layer implementation, the model started with poor performance but then gradually increased in accuracy. Around epochs 24-28 the model achieved its peak validation accuracy with %37, total training time was 26.54 minutes and the model used a significant memory with 15 GB, total integrated GPU usage was 1084.17 GPU seconds and total integrated power consumption was 225,517 Joules

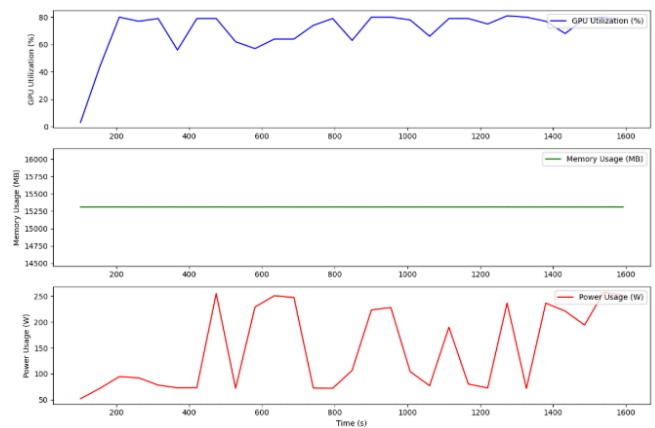

Figure 21. Third Implementation of FFT Layer

As discussed in the design choices section, we first utilized the Walsh-Hadamard Transformation (WHT) layer as a preprocessing step within our ResNet50 model. This approach enables the model to reduce redundancy in the input data, extract critical information and patterns from the image, and enhance generalization capability. The test concluded in 313.66 seconds, with GPU usage of 126.06 seconds, and a total power consumption of 28,164.15 Joules. The testing accuracy was similar to the base model benchmark scores, achieving a value of 64.23% (testing loss: 1.5571).

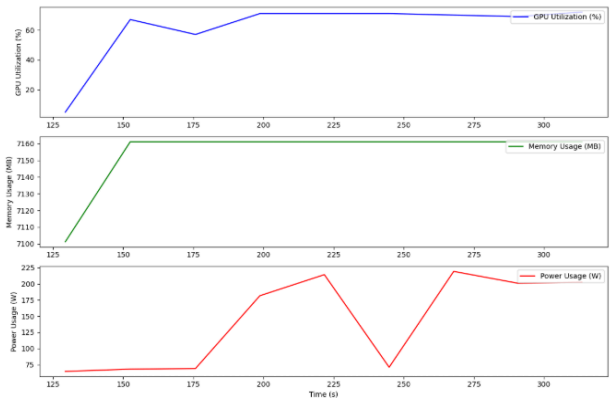

Figure 22. First Implementation of WHT Layer

After the first implementation, we followed up by placing the WHT layer after the first convolutional block of the ResNet50 architecture. The first convolutional blocks in the model capture basic spatial and low-level features; applying our signal processing transformation after this step might help capture critical patterns in the frequency domain. With this modification, we observed several noteworthy improvements in our training process.

Although the same parameters were used as in the earlier models, the total training time decreased to 270.11 seconds, and total integrated GPU usage slightly increased to 128.75 GPU-seconds from 126.06 GPU-seconds. This increase is expected, as in the previous model, the weights were locked, which reduced power consumption. Correspondingly, total integrated power consumption also increased to 31,761.43 Joules, which aligns with the architectural changes.

On the other hand, a critical improvement was observed in accuracy: the test accuracy increased to 73.46% (testing loss = 1.0556). With this enhanced accuracy, our model ranked as the 159th best model globally (including all models using the CIFAR-100 dataset) and the 18th best ResNet model in the CIFAR-100 benchmark scores [28].

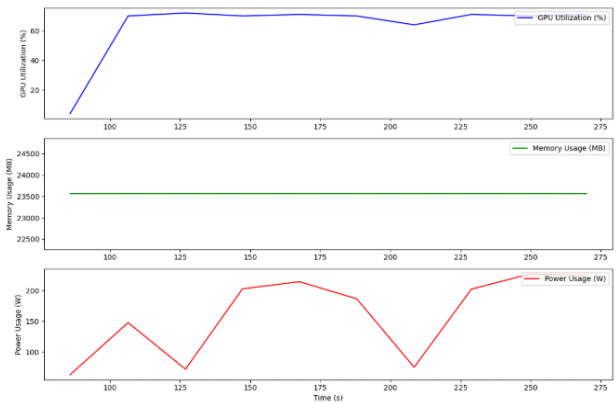

Figure 23. First Implementation of WHT Layer

Although this next integration wasn't implemented in other signal processing transformation architectures, we wanted to further evaluate the model by adding a dropout layer with a 0.5 dropout percentage and a regularizer. After testing, we observed the total training time as 220 seconds, total integrated GPU usage as 88.64 GPU-seconds, and total integrated power consumption as 20,057.10 Joules. The testing accuracy was 72.02%, with a test loss of 0.9839.

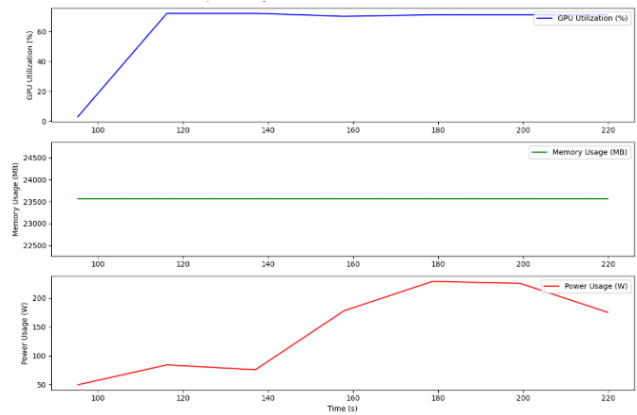

Figure 24. Second Implementation of WHT Layer (volume 2)

In the third implementation of WHT, After the training, we observed the total training time as 279.40 seconds, total integrated GPU usage as 113.93 GPU-seconds, and total integrated power consumption as 26433.78 Joules. However we observed a big improvement in testing accuracy with 78.62% ( testing loss = 0.8497) with these improvements, according to the benchmarks our new model performed as the 11[th] best ResNet architecture and 134[th] best model that utilizes CIFAR-100 dataset in the world.

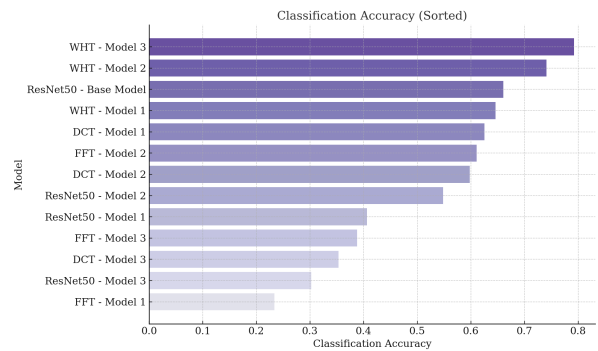

Figure 25. Model Accuracy Graph

| Model | Testing Accuracy | GPU Usage (GPU-sec) | Memory Usage (MB) | Power Consumption (KiloJoules) | Training Time (sec) |
|---|---|---|---|---|---|
| ResNet50 - Base Model | 0.6608 | N/A | N/A | N/A | N/A |
| ResNet50 - Model 1 | 0.4063 | N/A | N/A | 25605.19 | N/A |
| ResNet50 - Model 2 | 0.5482 | N/A5482 | N/A | 25607.19 | N/A |
| ResNet50 - Model 3 | 0.3024 | N/A | N/A | 25617.19 | N/A |
| WHT - Model 1 | 0.6460 | 126.06 | N/A | 28.164 | 313.66 |
| WHT - Model 2 | 0.7408 | 128.75 | 23561 | 31.761 | 270.11 |
| WHT - Model 3 | 0.7925 | 113.93 | N/A | 26.433 | 279.40 |
| DCT - Model 1 | 0.6256 | 285.68 | 23561 | 63.94 | 474.00 |
| DCT - Model 2 | 0.5977 | 341.23 | 23561 | 84.39 | 525.0 |
| DCT - Model 3 | 0.3531 | 395.72 | 23561 | 105.09 | 604 |
| FFT - Model 1 | 0.2334 | 379.92 | 15301 | 79.138 | 945.41 |
| FFT - Model 2 | 0.6107 | 427.20 | 25603 | 101.254 | 685.38 |
| FFT - Model 3 | 0.3878 | 1143.50 | 25651 | 219.188 | 1506.98 |

Figure 26. Table of Results

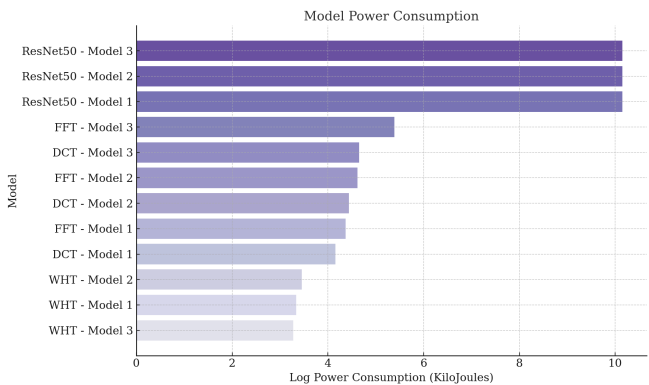

Figure 27. Model Power Consumption

## XI. DISCUSSION

Although the baseline model performed better in accuracy when compared to the first design implementation of our Walsh-Hadamard Transformation (WHT) Layer, the second and third design implementations of the WHT Layer greatly increased overall performance. When we used the second design implementation, our model ranked 17th in benchmark scores. With the third implementation, it improved to 12th place, below ResNet56 with non-convex learning by implementing the replica exchange stochastic gradient descent algorithm. By implementing WHT in both the early and later stages of the ResNet50 architecture, we propose a novel methodology that has the potential to inspire similar transformations in other domains for use in Convolutional Neural Network architectures, ultimately contributing to more power-efficient and improved deep learning models.

We observed that the models could extract high-frequency details such as edges and textures, as well as low-frequency components such as shapes and lines, with improved accuracy and reduced power consumption. Unlike other transformations, WHT operates with O(N log N) computational complexity, making the system more resource-efficient. Additionally, while implementing the WHT layer at various places in the architecture, we did not observe an increase in memory usage, which indicates that these transformations do not require additional memory consumption. When comparing these signal processing techniques with the baseline model and each other, we observed the following results. The ResNet50 base model performed better than variations of ResNet50 models. Although accuracy dropped in some cases, generalization improved, reducing the overfitting problem in the new models, which we will discuss later. WHT Model 3 outperformed all other models with an accuracy of 0.7925, surpassing even the ResNet50 baseline model.

Overall, the WHT variants achieved higher accuracy than the variants of DCT/FFT models. Our first design implementation of DCT showed decent performance with 62.56% accuracy but was still outperformed by the baseline model. Unlike WHT, adding 2D DCT to different places in the base ResNet50 model reduced performance. FFT models were more varied; the second FFT model was close to the first DCT model and outperformed all other FFT implementations. In conclusion, FFT models were outperformed by the WHT variants and the ResNet50 baseline model.

Since previous papers [26, 27] utilized ResNet50 as their primary base model, we also implemented this complex model. At the beginning of the project, we used MNIST and CIFAR-10 datasets to evaluate the discussed signal processing algorithms. However, due to the complexity of the ResNet50 model, we couldn't prevent overfitting. Even using multiple techniques such as adding L2 regularization (increased to 1e-3), data augmentation, unfreezing and freezing layers, and transfer learning did not reduce overfitting to a level where meaningful evaluations could be observed. Consequently, we used CIFAR-100 as our primary dataset to evaluate results and benchmark them against online results [28]. In the first design iteration, ResNet50 pre-trained on the "ImageNet" dataset was used without data augmentation, regularizers, or dropout layers. This gave us initial insights into ResNet50's performance with CIFAR-100. As shown in Figure 14, heavy overfitting was observed after the 13th epoch. Training accuracy increased, but validation accuracy decreased. Despite this, we achieved an accuracy of 0.6618 with a loss of 1.5186, similar to previous benchmark results [28].

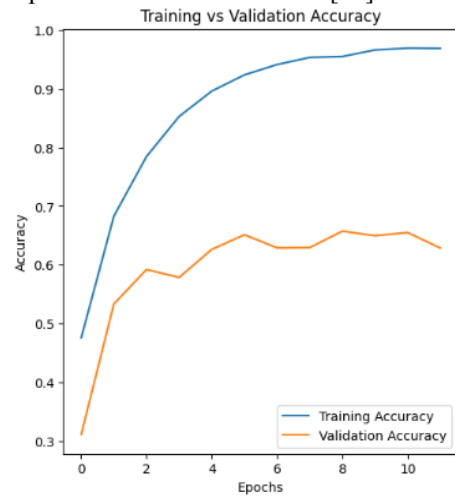

Figure 28. Training vs Validation Accuracy First Design Choice

On the test data, accuracy remained stable at 0.6608 with a test loss of 1.5223. We then introduced data augmentation, applying horizontal flips, rotations, zooms, contrast adjustments, random brightness, translations, and cropping. However, these augmentations also introduced noise, making training more challenging even for the ResNet50 model. Additionally, we added a dropout layer with a 0.4 probability, which, being higher than usual, caused the model to lose important features during training. By the 8th epoch, training accuracy was 0.5046 (test loss = 1.8868), and validation accuracy was 0.4063 (validation loss = 2.9524), as shown in Figure 29. Data augmentation techniques could not prevent overfitting.

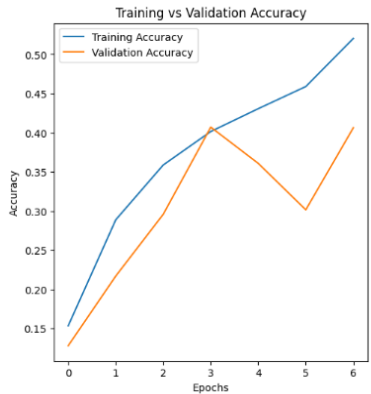

Figure 29. Training vs Validation Accuracy Second Design Choice

As referenced in [33], we wanted to use the idea of freezing and unfreezing layers in the ResNet50 model. By doing that, we can make some of the layers stop updating the weights during backpropagation, which can help reduce overfitting. Even using A100 GPUs for testing, the training took 45 epochs (1 hour and 30 minutes) to conclude. At the end, we got a training accuracy of 0.3725 (training loss = 2.5703) but with a validation accuracy of 0.5482 (validation loss = 1.8272), as can be seen in Figure 30. The difference between the validation and training accuracy can be linked to the existence of the dropout layers, which at one point made the model harder to learn new features. However, when we look at Figure 30, we can see that we successfully reduced overfitting with a slight decrease in overall accuracy. When we look at the benchmarks [28], base model ResNet50 implementations without adding any custom layers haven't addressed the problem of overfitting.

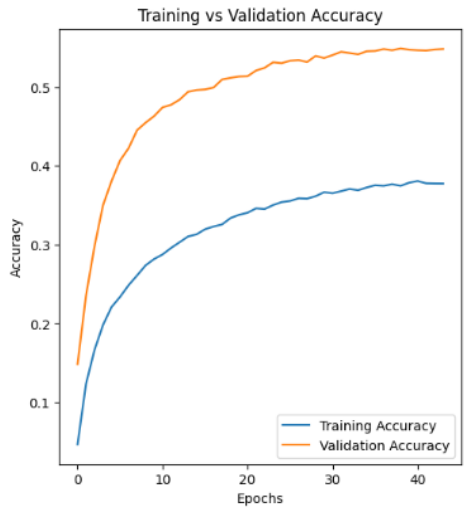

Figure 30. Training vs Validation Accuracy Second Design Choice

We concluded the training in two phases. In the first phase, we set a low-value learning rate (1e-4) with freezing of the ResNet50 base model architecture. In the second phase, we unfroze the last 50 layers of the ResNet50 model and further lowered the learning rate (to 1e-5) from 1e-4 to extract the features better. Apart from these changes, we haven't changed the other hyperparameters. Even though these implementations, in theory, make the data less complex, because of the complexity of our base dataset (CIFAR-100), the last implementation lowered the accuracies and made the model learn even harder. We will discuss the evaluation scores in the Evaluations section of this paper; training and validation accuracy can be observed in Figure 31.

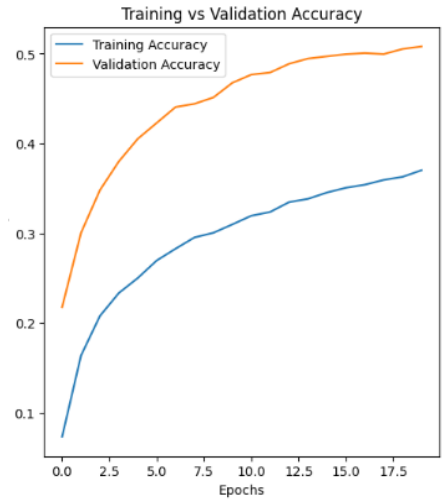

Figure 31. Training vs Validation Accuracy Second Design Choice

After these improvements, we managed to lower the risk of overfitting by adding various optimization, regularization, and other deep learning techniques. To observe the overfitting problem, we used several evaluation metrics. By observing the training accuracy, we can get a rough idea of the training process, observe overfitting in the model, and identify if the system starts to memorize some of the features. In some cases, we have saved time by correctly identifying the problem solely based on training and validation accuracy data during training. Validation accuracy, on the other hand, uses the same logic as training accuracy but instead of using the training dataset, it utilizes the validation dataset. Because of this, we can observe the model's performance on unseen data and evaluate its generalization performance.

Another evaluation metric we used throughout this project was the confusion matrix. As it is generally used for binary and multiclass predictions [10], we decided to use it for our CIFAR-100 dataset, which consists of 100 classes. By only implementing the confusion matrix, we can derive many evaluation metrics from it. For simplicity, we will explain this by focusing on binary classification problems, but the process is the same for multiclass classification problems. Simply, the confusion matrix consists of four regions: two rows corresponding to predicted values (predicted positive and negative, respectively) and two columns corresponding to the actual values (actual positive and negative, respectively). True Positive (TP) indicates that the model successfully predicts the positive (1) class, while True Negative (TN) indicates that the model successfully predicts the negative (0) class. False Positive (FP), on the other hand, indicates that the model was unsuccessful in predicting the actual positive class, which often refers to a Type I error. Finally, False Negative (FN) occurs when the model incorrectly predicts a negative class, which refers to a Type II error. In our case, for multiclass classification problems, this 2x2 matrix is expanded into an NxN table where N is the number of classes in the dataset. In this paper, we will generate 100x100 confusion matrices while working on the CIFAR-100 dataset. Using TP, TN, FP, and FN, we can derive the following performance metrics:

$$\text{Accuracy} = \frac{TP + TN}{TP + TN + FP + FN} \quad (15)$$

$$\text{Precision} = \frac{TP}{TP + FP} \quad (16)$$

$$\text{Recall} = \frac{TP}{TP + FN} \quad (17)$$

$$\text{F1 Score} = 2 \cdot \frac{\text{Precision} \cdot \text{Recall}}{\text{Precision} + \text{Recall}} \quad (18)$$

$$\text{Specificity} = \frac{TN}{TN + FP} \quad (19)$$

As can be seen in the formulations above, accuracy indicates the overall percentage of correctly identified predictions, precision indicates the true value of positive instances in predicted positive instances (positive predictive value), recall indicates the correctly identified positive instances (true positive rate), F1 score is the harmonic mean of precision and recall, and specificity can be observed as the true negative rate. Additionally, to provide better visualization, we also implemented a custom Grad-CAM Heatmap Overlay, which was investigated in detail in [34]. The main idea behind Grad-CAM is to visualize the areas that contribute the most strongly by overlaying a heatmap, which helps us understand the strong features solely based on the figure.

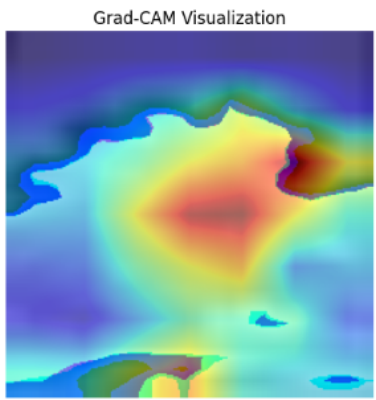

Figure 32. Grad-CAM Visualization

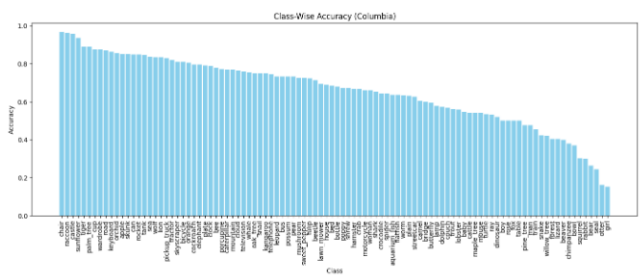

Figure 33. Class-Wise Accuracy Plot

Comparing the results of our models to the papers we had researched, we were unable to recreate the same level of accuracy. Two of the papers, "Deep Neural Network with Walsh-Hadamard Transform Layer For Ember Detection during a Wildfire" and "Block Walsh–Hadamard Transform-based Binary Layers in Deep Neural Networks", were able to achieve an accuracy of over 95% with their WHT implementations. However, this was to be expected. While we took great inspiration from these papers, we did not strive to recreate them, thus we could not produce the same specialized results. On the other hand, we were able to achieve the same conclusion as these research papers: implementing WHT in a neural network can increase the accuracy of the model more than other transforms, like DCT and FFT [7][9].

### XII. FUTURE WORK

There are numerous opportunities to improve and expand upon our research.

Ideally, each test would have been run multiple times, with results averaged to ensure reliability. However, due to limitations in computational resources and time constraints, the number of runs was restricted, particularly during the creation of the base model.

Our study focused on testing one neural network and one dataset, but future research could explore less complex ResNet models or other architectures. Additionally, examining and tuning the hyperparameters of the neural network could lead to significant performance improvements. Further optimization of each model would also be beneficial.

Expanding the scope of datasets and models presents another area for future work. For instance, datasets like Celeb-HQ and neural networks such as MobileNet could be explored. Implementing less complex models could also offer a valuable perspective. Beyond image data, it would be intriguing to apply these methods to other data types, broadening the research's applicability.

One challenge we encountered during initial testing was model overfitting. Developing more customized models with fewer layers and enhanced regularization techniques could yield better results in terms of both accuracy and GPU performance. Additionally, exploring the Fast Walsh-Hadamard Transform (FWHT), a more efficient variant of the Walsh-Hadamard Transform (WHT), could provide valuable insights. It would be interesting to investigate how FWHT impacts model accuracy and GPU performance efficiency.

### XIII. CONCLUSION

This paper examines the performance of three different applications of signal processing transformations in convolutional neural networks. The transformations we chose to analyze were the Walsh-Hadamard Transform, the Discrete Cosine Transform, and the Fast Fourier Transform. Using the pre-built ResNet50 model as a baseline with the CIFAR-100 dataset, we created multiple implementations of the transformation layers, comparing their accuracy performance as well as their computational performance. FFT performed the worst in terms of accuracy, achieving 0.6107 using our second model implementation. DCT performed slightly better, achieving an accuracy of 0.6256 using our first model implementation. WHT performed extremely well, achieving an accuracy of 0.7925. In terms of computational performance, we observed a similar trend. FFT had the least efficient performance with a longer GPU time and larger power consumption compared to the other transform implementations. DCT performed better than FFT, but not as well as WHT. The WHT model implementations consistently outperformed the other two transform implementations in test accuracy, GPU time, and power consumption. Overall, we were able to achieve our goal of increasing the accuracy of the model using WHT, achieving the eleventh-best model that fitted for CIFAR-100. Due to the complexity of ResNet50, we were unable to fully optimize the model due to a significant increase in the necessary computational power. Further research would require increased time to retrain ResNet50. Alternatively, a less complex model could provide the necessary solution to overfitting and increase the performance of the WHT implementation further.

## XIV. ACKNOWLEDGEMENT

We would like to thank Professor Zoran Kostic of Columbia University's Electrical Engineering Department and the TAs of ECBM 4040 for providing us with knowledge, tools, and the platform to be able to perform this research as a part of the course curriculum.

## XV. REFERENCES

[1] K. Hornik, "Approximation capabilities of multilayer feedforward networks," *Neural Networks*, vol. 4, no. 2, pp. 251–257, 1991.

[2] W. Fedus, B. Zoph, and N. Shazeer, "Switch Transformers: Scaling to Trillion Parameter Models with Simple and Efficient Sparsity," *Journal of Machine Learning Research*, vol. 23, no. 120, pp. 1–39, 2022.

[3] R. Quinn, "Microsoft invests in solar farms and nuclear power to support AI operations," *Politico*, 2024.

[4] Adobe Inc., "Filter basics," *Adobe Photoshop Help*, [Online].Available:https://helpx.adobe.com/photoshop/using/filter-basics.html.

[5] W. K. Pratt and N. P. Ross, "Walsh-Hadamard Transform Coding for Mariner 9 Pictures," NASA Jet Propulsion Laboratory, Pasadena, CA, Technical Report 32-1525, May 1971.

[6] Q. V. Le, T. Sarlos, and A. J. Smola, "Fastfood: Approximating kernel expansions in loglinear time," in *Proc. 30th Int. Conf. Mach. Learn. (ICML)*, Atlanta, GA, USA, Jun. 2013, pp. 244–252.

[7] H. Pan et al., "Deep Neural Network with Walsh-Hadamard Transform Layer For Ember Detection during a Wildfire", 2022 IEEE/CVF Conference on Computer Vision and Pattern Recognition Workshops (CVPRW), pp. 256–265, Jun. 2022. doi:10.1109/cvprw56347.2022.00040

[8] J. Park and S. Lee, "Energy-Efficient Image Processing Using Binary Neural Networks with Hadamard Transform", Lecture Notes in Computer Science, pp. 512–526, 2023. doi:10.1007/978-3-031-26348-4_30

[9] H. Pan, D. Badawi, and A. E. Cetin, "Block Walsh–Hadamard Transform-based Binary Layers in Deep Neural Networks", ACM Transactions on Embedded Computing Systems, vol. 21, no. 6, pp. 1–25, Oct. 2022. doi:10.1145/3510026

[10] J. W. Cooley and J. W. Tukey, "An algorithm for the machine calculation of complex Fourier series," *Math. Comput.*, vol. 19, no. 90, pp. 297–301, Apr. 1965.

[11] J. B. J. Fourier, *Théorie Analytique de la Chaleur*, Paris, France: Firmin Didot, 1822.

[12] A. V. Oppenheim and R. W. Schafer, *Discrete-Time Signal Processing*, 3rd ed. Upper Saddle River, NJ, USA: Prentice Hall, 2010.

[13] N. Ahmed, T. Natarajan, and K. R. Rao, "Discrete cosine transform," *IEEE Transactions on Computers*, vol. C-23, no. 1, pp. 90–93, Jan. 1974.

[14] ISO/IEC 11172-2:1993, *Information technology—Coding of moving pictures and associated audio for digital storage media at up to about 1,5 Mbit/s—Part 2: Video (MPEG-1 Video)*, 1993; ISO/IEC 13818-2:1995, *Information technology—Generic coding of moving pictures and associated audio information—Part 2: Video (MPEG-2 Video)*, 1995; ITU-T Recommendation H.263, *Video coding for low bit rate communication*, Feb. 1998.

[15] N. Ahmed, "How I proposed the discrete cosine transform to the NSF," *IEEE Signal Processing Magazine*, vol. 21, no. 1, pp. 16–17, Jan. 2004.

[16] K. R. Rao and P. Yip, *Discrete Cosine Transform: Algorithms, Advantages, Applications*, Academic Press, 1990.

[17] A. K. Jain, *Fundamentals of Digital Image Processing*, Prentice Hall, 1989.

[18] *TensorFlow Core v2.12.0: tf.signal.dct*, TensorFlow Documentation. [Online]. Available: https://www.tensorflow.org/api_docs/python/tf/signal/dct.

[19] W. H. Press, S. A. Teukolsky, W. T. Vetterling, and B. P. Flannery, *Numerical Recipes in C: The Art of Scientific Computing*, 2nd ed., Cambridge University Press, 1992.

[20] J. G. Proakis and D. G. Manolakis, *Digital Signal Processing: Principles, Algorithms, and Applications*, 4th ed. Upper Saddle River, NJ, USA: Prentice Hall, 2007.

[21] J. J. O'Connor and E. F. Robertson, "Jacques Salomon Hadamard," MacTutor, https://mathshistory.st-andrews.ac.uk/Biographies/Hadamard/.

[22] J. J. O'Connor and E. F. Robertson, "Joseph Leonard Walsh", MacTutor, https://mathshistory.st-andrews.ac.uk/Biographies/Walsh_Joseph/.

[23] N. Ahmed, "Walsh-Hadamard Transform based interframe transform coding," NASA Ames Research Center, Moffett Field, CA, Technical Report, 1972.

[24] J. J. Sylvester, "Thoughts on inverse orthogonal matrices, simultaneous sign successions in a series of linear inequalities, and tessellated pavements in two or more colours, with applications to Newton's rule, ornamental tile-work, and the theory of numbers," *Philos. Mag.*, vol. 34, no. 232, pp. 461–475, 1867.

[25] K. He, X. Zhang, S. Ren, and J. Sun, "Deep Residual Learning for Image Recognition," in *Proc. IEEE Conf. Comput. Vis. Pattern Recognit. (CVPR)*, Las Vegas, NV, USA, Jun. 2016, pp. 770–778.

[26] A. Patel and A. Mehta, "Integration of Discrete Cosine Transform in Convolutional Neural Networks for Improved Generalization," *arXiv preprint arXiv:2103.05645*, 2021. [Online]. Available: https://arxiv.org/abs/2103.05645

[27] X. Xu, Y. Xu, and J. Han, "Frequency Domain Neural Network for Efficient Learning and Inference," *arXiv preprint arXiv:2005.05673*, 2020. [Online]. Available: https://arxiv.org/abs/2005.05673

[28] "Image Classification on CIFAR-100," *Papers With Code*, [Online]. Available: https://paperswithcode.com/sota/image-classification-on-cifar-100?tag_filter=3.

[29] Guo, Q., Wu, Z., Chen, Z., & Huang, J. (2020). *Spectral Analysis Network for Evaluating Feature Frequency*. IEEE Transactions on Neural Networks and Learning Systems, 31(11), 4536-4549.

[30] Kundu, S., Das, R., & Pal, S. (2019). *Enhancing Convolutional Neural Networks with Frequency Domain Features for Image Classification*. Neural Processing Letters, 50(3), 2135-2151.

[31] Wang, L., Sun, Y., Zhang, Y., & Duan, Y. (2018). *Efficient Convolutional Neural Networks with Discrete Cosine Transform for Mobile Applications*. Proceedings of the IEEE/CVF Conference on Computer Vision and Pattern Recognition (CVPR).

[32] A. V. Oppenheim, A. S. Willsky, and I. T. Young, *Signals and Systems*, 2nd ed., Prentice Hall, 1997.

[33] J. Yosinski, J. Clune, Y. Bengio, and H. Lipson, "How transferable are features in deep neural networks?," in *Advances in Neural Information Processing Systems*, vol. 27, pp. 3320–3328, 2014.

[34] R. R. Selvaraju, M. Cogswell, A. Das, R. Vedantam, D. Parikh, and D. Batra, "Grad-CAM: Visual Explanations from Deep Networks via Gradient-Based Localization," in *Proc. IEEE Int. Conf. Comput. Vis. (ICCV)*, Venice, Italy, Oct. 2017, pp. 618–626.

[35] A. Aslam and S. Farhan, "Enhancing rice yield prediction: a deep fusion model integrating ResNet50-LSTM with multi source data," PeerJ Computer Science, vol. 10, Aug. 2024. doi:10.7717/peerj-cs.2219

[36] Link to github: https://github.com/ecbme4040/e4040-2024fall-project-pbdy-by2385-dyh2111-pad217

XVI. APPENDIX

| UNI | by2385 | dyh2111 | pad2176 |
|---|---|---|---|
| Last Name | Yilmaz | Harvey | Dhuri |
| Fraction of (useful) total contribution | 1/3 | 1/3 | 1/3 |
| Coding | 1/3 | 1/3 | 1/3 |
| Report | 1/3 | 1/3 | 1/3 |
| Research | 1/3 | 1/3 | 1/3 |

Google drive folder for the best model : *https://drive.google.com/drive/folders/1RfDSWnZ5p8gFUOkPu16zDH8_GTo7-ndB?usp=drive_link*

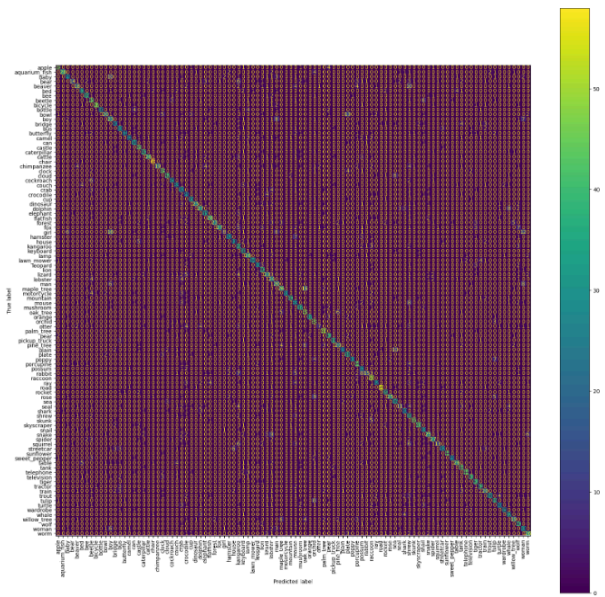

Figure A1. Confusion matrix for the baseline model ResNet-50, red indicates positive correlation where blue indicates the negative correlation

Link to Github:
https://github.com/ecbme4040/e4040-2024fall-project-pbdy-by2385-dyh2111-pad2176

**Algorithm 4** GPU Monitor Callback
---
**Require:** GPU with NVML support
**Ensure:** Monitoring data logged
  Initialize:
    GPU utilization (%)
    Memory usage (MB)
    Power usage (Watts)
    Timestamps (elapsed times)
    Start time and GPU handle
  Begin epoch:
    Start tracking time
    Initialize NVML and GPU handle
  In each epoch:
    Update elapsed time
    Record GPU utilization, memory, and power
    Convert metrics to appropriate units
    Append data to logs
    Print metrics
  Shutdown NVML after all epochs

**Algorithm 5** Inverse Walsh-Hadamard Transform
---
**Require:** Input image $X$ of size $n \times n$, $n$ is a power of 2
**Ensure:** Output matrix after applying Inverse Walsh-Hadamard Transform
  Generate Walsh-Hadamard matrix $H$ of size $n \times n$
  Ensure the input image $X$ is of type `float32`
  Apply the Inverse Walsh-Hadamard Transform:
    $X \leftarrow H \cdot X$     ▷ Multiply the Walsh-Hadamard matrix with the image
    $H^T \leftarrow \text{Transpose}(H)$ ▷ Get the transpose of the Walsh-Hadamard matrix
    $X \leftarrow X \cdot H^T$     ▷ Multiply the result with the transpose
  Normalize the result: $X \leftarrow X/n^2$
  **return** $X$

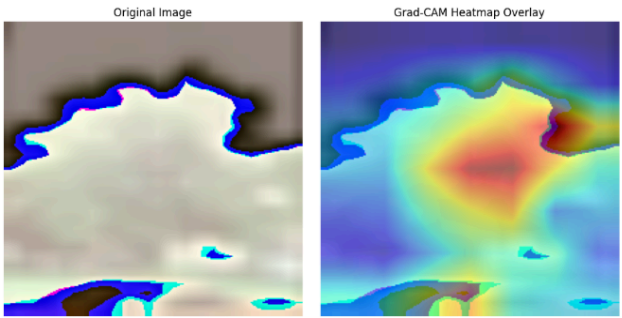

A1. Original image and a transformed image that GradCAM Heatmap applied on

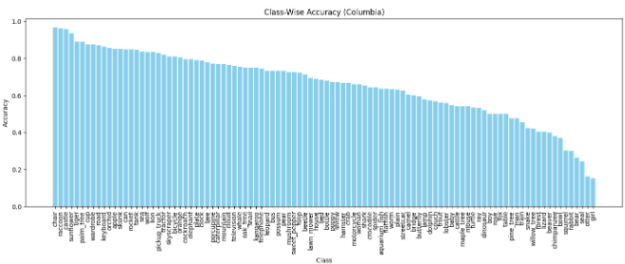

A2. Class wise accuracy of predictions histogram of the baseline ResNet-50 model

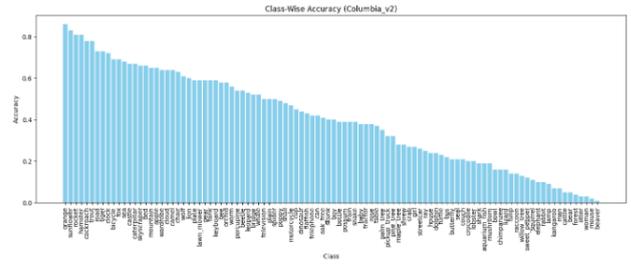

A3. Class wise accuracy accuracy of predictions histogram of the second ResNet-50 model

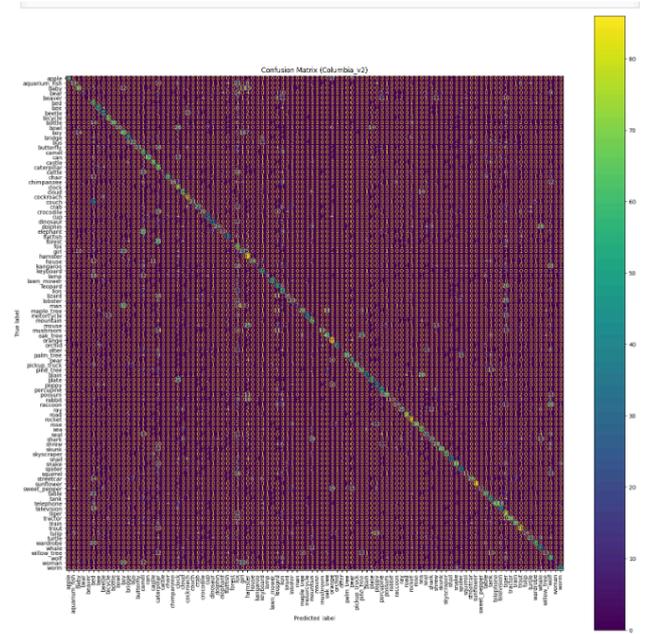

A4. Confusion matrix for the baseline model 2 of ResNet-50, red indicates positive correlation where blue indicates the negative correlation

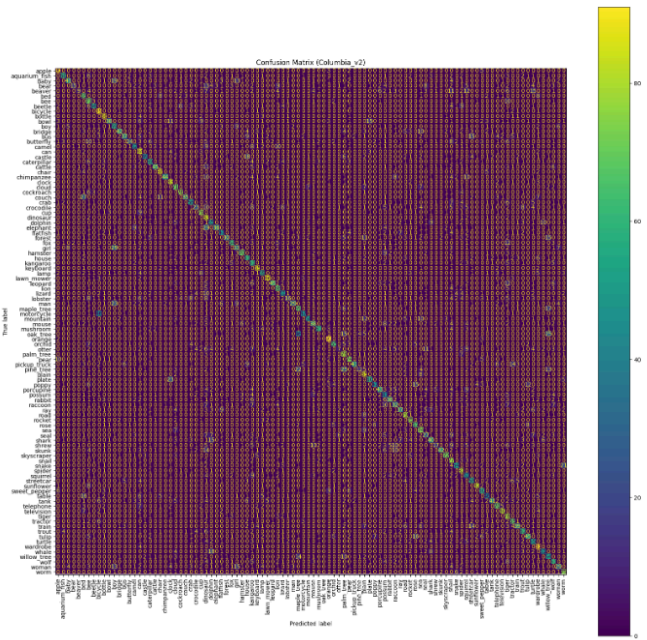

A5. Confusion matrix for the baseline model 3 of ResNet-50, red indicates positive correlation where blue indicates the negative correlation

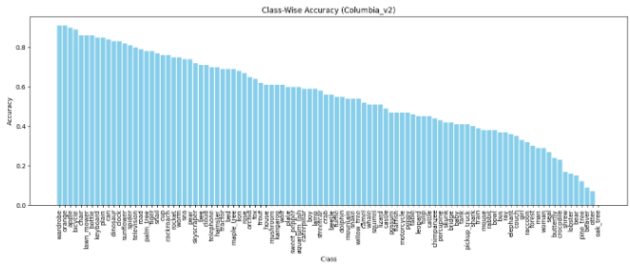

A6. Class wise accuracy accuracy of predictions histogram of the third ResNet-50 model

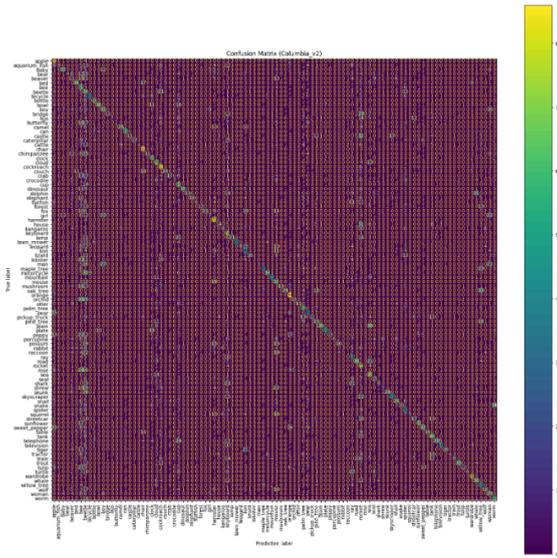

A7. Confusion matrix for the baseline model 4 of ResNet-50, red indicates positive correlation where blue indicates the negative correlation

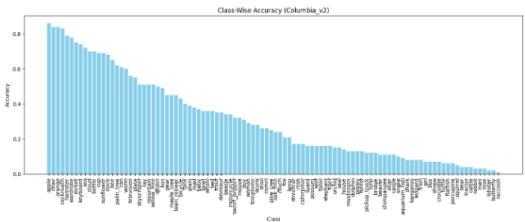

A8. Class wise accuracy accuracy of predictions histogram of the third ResNet-50 model

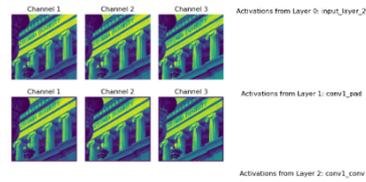

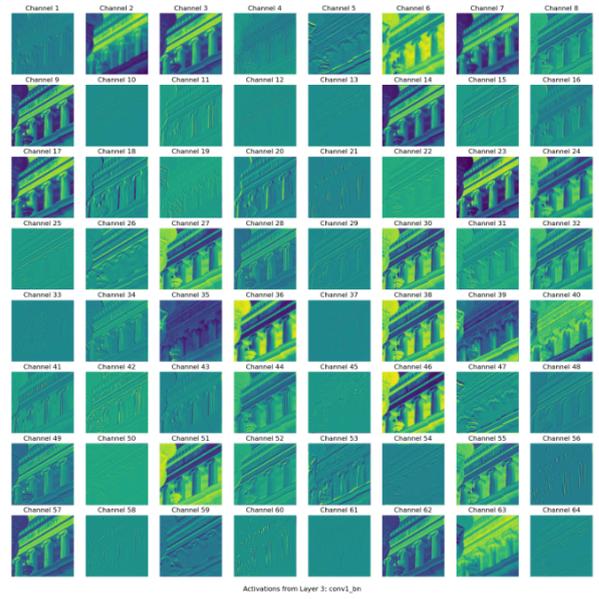

A9. ResNet-50 base model visualization of the featuremaps, early layers

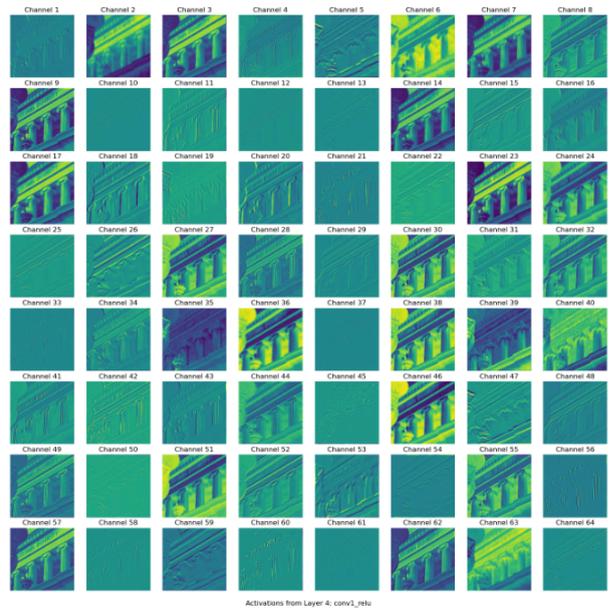

A10. ResNet-50 base model visualization of the feature maps, middle layers

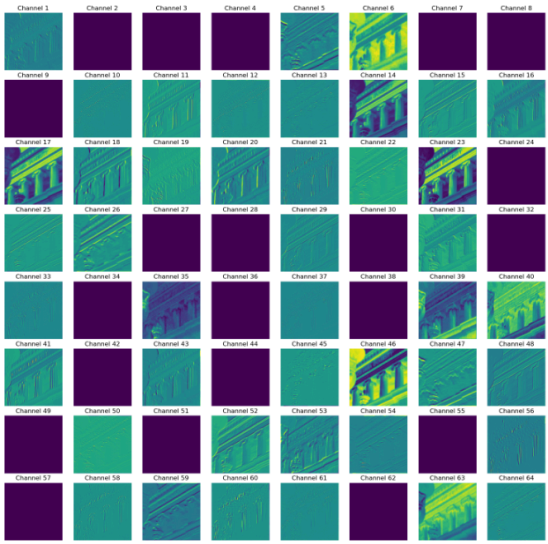

A11. ResNet-50 base model visualization of the featuremaps,last layers